# Enhancing Illicit Activity Detection using XAI: A Multimodal Graph-LLM Framework

Jack Nicholls
jack.nicholls@ucdconnect.ie
University College Dublin, School of
Computer Science,
Dublin, Ireland

Aditya Kuppa
aditya.kuppa@ucd.ie
University College Dublin, School of
Computer Science,
Dublin, Ireland

Nhien-An Le-Khac
an.le@ucd.ie
University College Dublin, School of
Computer Science,
Dublin, Ireland

## ABSTRACT

Financial cybercrime prevention is an increasing issue with many organisations and governments. As deep learning models have progressed to identify illicit activity on various financial and social networks, the explainability behind the model decisions has been lacklustre with the investigative analyst at the heart of any deep learning platform. In our paper, we present a state-of-the-art, novel multimodal proactive approach to addressing XAI in financial cybercrime detection.

We leverage a triad of deep learning models designed to distill essential representations from transaction sequencing, subgraph connectivity, and narrative generation to significantly streamline the analyst's investigative process. Our narrative generation proposal leverages LLM to ingest transaction details and output contextual narrative for an analyst to understand a transaction and its metadata much further.

## 1 INTRODUCTION

The evolving landscape of financial cybercrime has caused a seismic shift in the focus of industry and academic research to prevent fraudulent activity [9]. Never has there been a time where people are more susceptible to cybercriminal victimisation including scamming campaigns through SMS (smishing), phone calls (phishing), video calls (vishing), emails (business email compromise), or being directly contacted through messaging services built into social media applications. This has been impacted even more with the aid of malicious Large Language Models (LLMs) like FraudGPT [8].

Deep learning and machine learning modelling has emerged as a strong tool to aid multiple departments of business institutions including financial crime, cybersecurity, compliance, risk, treasury, and more. A key aspect for any financial institute engaging in anti-money laundering (AML) and financial crime compliance includes the explainability behind the identification of illicit transactions or behaviour on their networks.

We aim to address the investigative process of financial cybercrime through a novel, multimodal, and state-of-the-art deep learning pipeline using LLMs, Generative Pre-Trained (GPT) models, and Graph Neural Networks (GNNs). We take advantage of the capabilities of GPT models to capture the sequential behaviour of transactions through generating embeddings. We then couple this with two different models to enhance the illicit activity investigation and detection through use of narrative generation using LLMs, and transaction subgraph connectivity identification using GNNs.

These techniques present a proactive, multimodal methodology to aid analysts in their investigative duties including advanced explainability. The architecture includes the creation of transaction narratives which translate the features of a transaction into statements which allow an analyst to understand and compare transactions through language rather than numerical or statistical comprehension alone. The proposed system allows any user to begin an investigation through simple natural language queries into a search interface. The multimodal architecture provides relevant answers such as malicious addresses or activity with explainability through similar examples of other identified subgraphs or transaction sequences.

Graphs excel at depicting the connections among data points within a dataset, offering a distinct advantage compared to conventional tabular data due to their innate clarity and versatility, particularly in terms of visualisation. Explanation in Graph Learning allows the comprehension of important subgraph structures where a small subset of node features can play a crucial role in Graph Neural Networks (GNNs) predictions. Our connectivity measurement connects the identified transaction sequences in a network to find similarly occurring events. For example, if a string of 25 transactions are identified in a cryptocurrency mixing pattern to obfuscate origin of illicit funds and this sequence is identified by our GPT. We can then create a subgraph on the entire network that captures the 25 transactions and our GNN will calculate an embedding. Future transactions and subsequent generated subgraphs from the transaction can be compared with a similarity score to our identified mixing sequence and subgraph. Our LLM narrative prompt will provide transaction details and the associated metadata with the queried or flagged transaction. The narrative prompt will be validated against a critic prompt, which is typically another LLM which will critique and review the coherence, relevance, accuracy, and completeness of the newly generated narrative statement prior to an analyst reviewing it.

This multimodal approach with XAI at the core of its architecture is to assist the investigative journey of analysts, starting at the entry to the senior level with a generative report which captures granular detail to aid in the downstream reporting requirements. Another avenue of generation is in the area of Suspicious Activity Reports (SARs), a regulatory requirement from financial institutions and various merchants when engaging in potentially illicit transactions or activity.

In section 2 we present the relevant background information. Then we describe the relevant work and the gaps we have identified to the best our knowledge in section 3. In section 4 we describe our methodology. Section 5 is our discussion of techniques. This includes a discussion on the direction and future work of our multimodal architecture and pipeline including human-in-the-loop review of transaction narratives. Section 6 is our concluding remarks.

## 2 BACKGROUND

The background section of our paper outlines financial cybercrime and the multiple avenues our framework can be applied to. We also cover background on graphs and their applications in tackling financial cybercrime and the associated graph deep learning algorithms that can be applied to the networks. We cover LLMs and how their embeddings are applied in our multimodal architecture.

### 2.1 Financial Cybercrime

Financial cybercrime has been defined as "a combination of financial crime, hacking, and social engineering committed over cyberspace for the sole purpose of illegal economic gain"[9]. It encapsulates a large breathe of financial crimes that take place over cyberspace, and are evolving rapidly with the newer avenues of extorting, scamming, and defrauding people with new technology like cryptocurrency enabling transfer of value without the requirement of a financial intermediary.

Machine learning methods have evolved not only to screen transactions of a financial institution to prevent illicit transactions from executing on their networks, but also monitor the behaviour of a user to anticipate anomalous activity to identify unusual behaviour indicative of illicit activity which can protect a customer further.

### 2.2 Graph Learning

Graphs, or networks, are relational datasets which have connections between data points. Graphs are represented as $G = V, E$, where $V$ is the set of nodes, and $E$ is the sets of edges or connections which captures the relationships between the nodes.

Graph deep learning (GDL) models, such as Graph Attention Networks (GATs) or Graph Convolutional Networks (GCNs) use the graph datasets as inputs to calculate embeddings or representations that capture structure and relationships between data points. They can be used in node classification tasks including transaction monitoring [9]. Our use of GDL is to calculate embeddings that capture subgraph structures on networks with transactions of interest. The embeddings will be extracted and concatenated with GPT embeddings to identify and discover similar structures based on transaction sequence embeddings.

### 2.3 Large Language Models

Large Language Models (LLMs) are highly scaled pre-trained language models (PLMs) with billions of more parameters increasing the capacity of downstream tasks [17]. LLMs are predominantly constructed using the Transformer architecture [14] wherein layers of multi-head attention are stacked in a deep neural network.

LLMs like ChatGPT by OpenAI have been a breakthrough in artificial intelligence modelling and have seen a wide amount of application in industry and research. A survey by Yang et al. [15] presented the various practical applications that people are applying LLMs to in their downstream natural language processing (NLP) tasks. These include: i) traditional language understanding include the basic tasks like text classification and named entity recognition (NER), ii) natural language generation is split into two categories which includes paragraph summarisation, and then "open-ended" generation like code creation, iii) knowledge-intensive tasks would include a closed-book question style examination, where normally a person would take this test with extensive memorisation of real-world knowledge would be required, iv) reasoning ability addresses the models ability to have 'commonsense' where a LLM needs to be able to retain factual knowledge but simultaneously perform several inference steps on that retained fact.

LLMs use text embeddings to make relations between words and structure sentences and are crucial to the ability to perform tasks of reasoning and generation. To leverage from the the use of LLMs for tasks outside of prepackaged LLMs suites like ChatGPT, we require access to the underlying embeddings generated in the models. DaVinci [11] is an example of a LLM that generates visible embeddings that will allow integration with other deep learning embeddings.

*2.3.1 Critic Prompts.* As we generate narratives to capture the context of a given transaction, we wish to validate these outputs. An emerging technique to validate or improve on a generative prompt is to pass it through another LLM [2, 6]. We deploy a similar methodology in our architecture to verify and validate our generated narratives to ensure that the context provided to the analyst is sound.

### 2.4 Combining Graph Deep Learning, Transformer and LLMs

The graph applications for LLMs are still at an early stage. Our contribution includes a novel method leveraging of LLMs and Pre Trained transformer models for practical use case in financial cybercrime investigation.

We leverage from several techniques to create our multimodal architecture. Our methodology can be broken up into three core areas: i) Transaction Sequence Embeddings, ii) Graph Embeddings for Connectivity, and iii) Narrative Generation for Contextualisation.

Embeddings provide powerful avenues for tabular deep learning application [3]. Models like GPTs and GNNs condense large, complex data structures like a cryptocurrency network down to numerical representations that can be used in discovery and inference of specific activity. Our paper aims to concatenate embeddings which capture multiple representations of a transaction network, and through similarity measurements discover potential illicit activity or relevant investigative information for an analyst. This is a novel approach to processing embeddings by multiple models. Many techniques include the use of embeddings as direct features in another deep learning network with the ambition of increasing evaluation metrics.

## 3 RELATED WORK

Research has been published in the domain of applying LLM to predicting output of graph datasets. Work by Chen et al. [1] explored the potential of LLMS in learning on Text Attributed Graphs (TAGs). They explain limitations on the shallow embeddings produced by Bag-of-Words [4] and Word2Vec [7] and their difficulty in processing polysemous words [12]. With the advent of LLMs and the breakthrough in products like ChatGPT, they posed two hypotheses for LLMs to tackle graph learning and more specifically node classification. The first was whether they could integrate LLM text embeddings with their data pipeline for GNN classification improvements, and the second was whether they could deploy a LLM as a predictor in a node classification task. They experimented on Cora, PubMed, OGBN-arXiv, and OGBN-products. Their experimentation revealed that enhancing their node features with LLM embeddings improves performance of the GNN classifiers.

There have been other breakthroughs in joint pre-training of different data types by coupling text deep learning images and video. CLIP [13] matches text encoding and image embedding in the same representation space [16] allows application in parsing video and images and extracting text features through computer vision. A survey by Zhang et al. [16] details the extensive applications of LLMs in various fields such as video, music, gaming, and graphs. LLMs application in the area of graphs is predominantly in the area of graph generation, rather than the use of LLMs in node classification demonstrated by Chen et al. [1].

To the best of our knowledge, the research is sparse on a multimodal holistic approach to tackling financial cybercrime with the implementation of transformers, LLMs, and GNNs.

## 4 METHODOLOGY

The XAI system proposed aims to assist analysts and auditors in their investigative tasks. Users can easily start their investigation by inputting natural language queries into a search interface. In response, the system provides relevant answers, malicious addresses with context descriptions, and transaction graphs supporting the query.

The methodology we adopt is multi-step. First, embeddings are generated to capture transaction sequences, connectivity, and their relevancy to external events. We leverage a Pre-trained Transfomer network which is trained specifically to capture the transaction sequences. To incorporate and contextualise information from external sources, a GPT-based zero-shot method is used to generate narratives from the events surrounding the transaction and its embeddings. To capture the connectivity of transactions and their neighbourhood, we train a graph transformer network, capturing the complexity of interactions and use the embeddings of each transaction at answering time. When an analyst poses a query, we search these embeddings to identify accounts exhibiting behaviours consistent with the question's intent. Additionally, we probe the context embeddings related to the account in question to gather relevant data. Finally, by searching through graph embeddings, we can pinpoint accounts that share similar transactional patterns. Figure 1 displays the multimodal approach to capture multiple layers of embeddings.

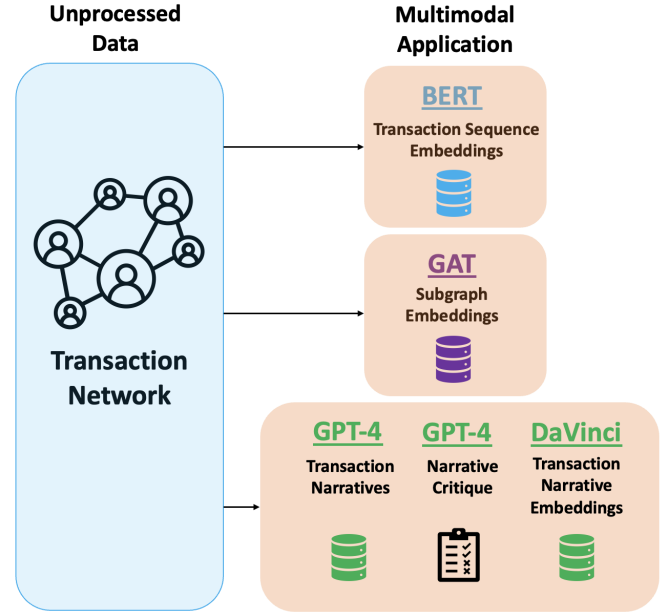

Figure 1: The unprocessed transactional network is the example used in this diagram which is then passed through a multimodal architecture extracting multiple layers of embeddings capturing different representations and relationships in the dataset. For each transaction in the dataset: BERT extracts transaction sequence embeddings, GAT will capture the subgraph embeddings, and GPT-4 will go through a cycle of transaction narrative contextual generation with narrative self-critiquing and then the final transaction narrative is converted using DaVinci to a visible embedding.

### 4.1 Embeddings

**Transaction Embeddings**: Given a transaction $T$, its embedding, $E_T$, captures its attributes. We employ a BERT-based model to generate these embeddings, allowing for high-dimensional sequential transaction representations. The model used is a fine-tuned architecture of BERT specific to support transaction sequences [5].

**Graph Embeddings for Connectivity**: The Ethereum transaction network is mapped to a graph $G$ where nodes represent transactions and edges symbolize their connectivity. We employ a GAT to produce embeddings $E_{G'}$ for a transaction graph involving an account that captures localized transactional patterns.

**Narrative Generation for Transaction Contextualisation**:

For each transaction $T$, an associated narrative $N$ provides insights in a digestible manner. Such narratives can assist analysts, auditors, or even regular users in understanding the context, significance, and potential ramifications of a transaction without delving into the technical details. First, we gather transaction details ($T$), including sender and receiver addresses, amount, date, and other metadata to generate narratives. The meta-information ($M$) includes the addresses involved, such as known associations, previous transaction behavior, etc. An external event data ($E$), such as significant price fluctuations, security breaches, or network upgrades, which may influence the transaction context, is integrated to understand the transaction context. For example, "On [Date], [Amount] was transferred from [Address A] to [Address B]. This transaction occurred during [External Event]. Notably, [Address A/B] has [Meta Information]." This narrative highlights the significance of the transaction, the account types involved, and the external event relevance. The details about $T$, $M$, and $E$ are given to the LLM and utilize its capability to craft a detailed narrative $N$ in a zero-shot manner.

> **Narrative prompt:** "Given the provided cryptocurrency transaction details, its associated metadata, and relevant external events, generate a concise 3-point narrative encapsulating the transaction's essence."
> {transaction_details}
> {meta_information}
> {external_events}

Upon generating a narrative $N$ for a transaction $T$, the critic mechanism evaluates $N$ against set criteria, producing feedback $F$ that helps in refining the narratives or flagging anomalies. The critic prompt will check for- (a) Coherence: Does the narrative flow logically? (b) Relevance: Does it highlight the most salient features of $T$?, (c) Accuracy: Does the narrative correctly represent $T$? (d)Completeness: Are all key details of $T$ captured? . The critic prompt improves the narrative based on the criteria once the $N$ and $T$ are submitted to the critic. This mechanism ensures that the narratives captured are of high quality and relevance.

> **Critic prompt:** "Review the provided narrative generated for the cryptocurrency transaction. Assess its coherence, relevance, accuracy, and completeness. Provide refined output as an improvement if necessary."
> {generated_narrative}
> {transaction_details}

Next, we concatenate transactions $E_T$ and narratives $E_N$ into representation, $E_C$. $E_C = f(E_T, E_N)$

### 4.2 Retrieval of Graph, Transaction Sequence, and Subgraphs Search

Given a query $Q$ from an analyst, the system performs a multi-layered retrieval process:

**Transaction and Narrative Embedding Search (Sequence):** Utilising the embedding $E_T$ generated for each transaction, the system performs a cosine similarity or nearest neighbor search to fetch transactions closely resembling the query. This stage ensures we capture transactions with inherent properties similar to the given query. With narrative embeddings in play, analysts can search based on raw transaction attributes and the nature, context, or intent of transactions represented by the narratives. For instance, an analyst could query for "transactions that indicate suspiciously large transfers in a short time," and the system could utilise $E_N$ to identify narratives (and thus transactions) that match this intent. Also, as narratives are reviewed, critiqued, and potentially corrected by analysts or using the critic prompt system, this feedback can be used to fine-tune the embedding models, ensuring that the representations become even more accurate and insightful overtime. This process gives higher flexibility and depth, allowing analysts to dive into the Ethereum data from multiple perspectives and with varied objectives.

$$\text{Similarity}(Q, T) = \frac{Q \cdot E_T}{\|Q\| \|E_T\|} \quad (1)$$

**Graph-based Retrieval (Connectivity):** Using graph embeddings, the system scans for nodes (accounts) that have interacted in manners similar to the queried transaction. This fetches the direct participants of similar transactions and their immediate neighbors in the graph, providing a richer context.

For a given node $v$ with embedding $E_v$, the similarity score with the query is computed, and nodes are ranked accordingly.

$$\text{similarity}(Q, T) = \frac{Q \cdot E_v}{\|Q\| \|E_v\|} \quad (2)$$

**Subgraph Extraction:** After extracting nodes of interest, the system extracts relevant subgraphs that contain the nodes and their direct neighborhoods. This gives a full view of the transaction environment, capturing patterns, cycles, and anomalies.

Combining transaction embeddings with graph-based techniques provides a comprehensive and contextual view of the queried transactions, aiding analysts in deeper exploration, anomaly detection, and insight extraction. Figure 2 displays the process of querying a transaction and retrieving a subgraph with supplemental contextual narrative.

## 5 DISCUSSION

The key function of this multimodal approach is to aid the investigative analyst in querying, labelling, and understanding transactions in a network based on the features available to them. The analyst investigation process begins at the first level, where a junior analyst (Level 1) will receive a flag or instigate an investigation on a suspicious or unusual transaction. Without sufficient experience, their judgment will be passed on to a higher, more knowledgeable, and/or experienced analyst. This process requires a summarisation of each subsequent analyst's opinions, reasoning, and evidence for why they have flagged or cannot make a sufficient judgment on a transaction. By leveraging our proposed method the flagged alert is contextualized with transaction details, a contextual narrative, and provides transactions of a similar nature to enhance the investigation. Additionally, tasks such as populating suspicious transaction reports (STRs) and other regulatory requirements can be expedited with the assistance of our method.

The proposed method has multiple advantages. For example, the multimodal approach can help identify mixing patterns, CoinJoins, and other obfuscation methods deployed by cybercriminals in the cryptocurrency space. Mixing patterns can include transactions from very distant points on the network, purposely performed to create a more challenging environment for analysts to investigate [10]. Figure 3 demonstrates the opportunity by having identified a mixing operation through the sequencing embedding stage, with the additional subgraph retrieval method it would be possible to identify other mixing patterns scattered across the entire blockchain rather than having to store the entire blockchain and all transactions in memory, as is the case for heuristics applied to the network.

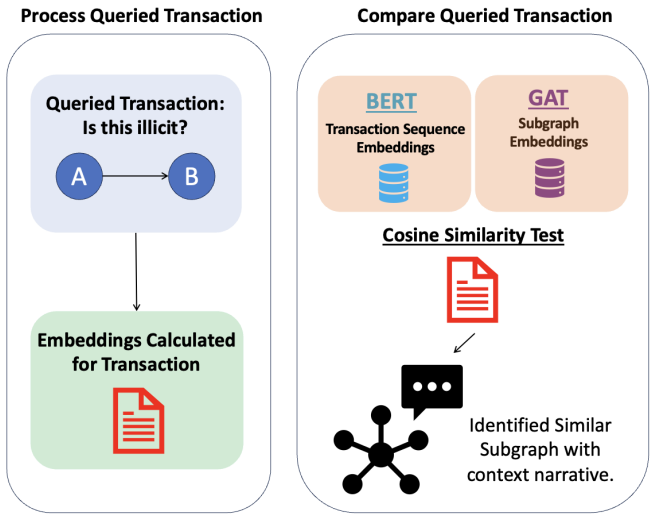

Figure 2: An analyst will query a particular transaction on the network such as "is this illicit?". The embeddings for the queried transaction are calculated. These embeddings are then compared with the originally processed embeddings by the BERT and GAT models. The cosine similarity method is used to search for other subgraphs that exhibit similar sequencing and features. Accompanying this will be a narrative providing context to the analyst answering their original query.

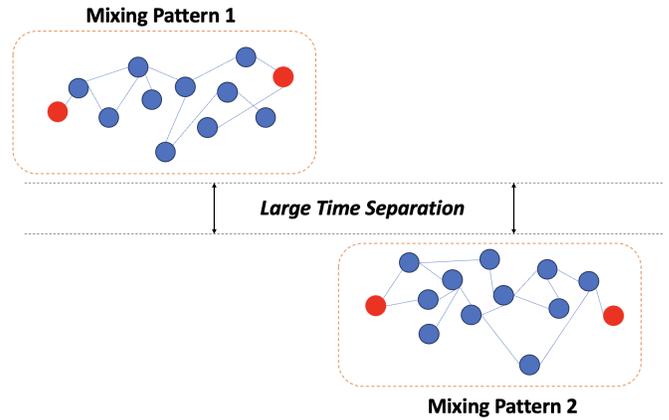

Figure 3: Mixing patterns are hard to manually follow through a visualisation tool like a transaction navigator. Mixing patterns are also scattered across the blockchain with large time separations between different patterns. This diagram shows a mixing pattern 1 on the top, separated by time against the second mixing pattern. The multimodal model has opportunity to leverage from the subgraph retrieval method to identify the second mixing pattern in the blockchain.

## 6 CONCLUSION

We introduced a multimodal strategy that harnesses the power of multiple deep learning models to offer a comprehensive and interpretable investigative toolkit for analysts. As the sophistication of financial cybercriminals increases, the experience and knowledge required from analysts increases dramatically. Our method not only alleviates these demands but also enhances the skillset and methodologies of analysts combatting financial crime. As a next step, we plan to conduct an extensive user study on the proposed method to assess its impact on enhancing the overall efficiency and productivity of analysts.

## ACKNOWLEDGMENT

This publication has emanated from research conducted with the financial support of Science Foundation Ireland under Grant number 18/CRT/6183. For the purpose of Open Access, the author has applied a CC BY public copyright license to any Author Accepted Manuscript version arising from this submission.

## REFERENCES
[1] Zhikai Chen, Haitao Mao, Hang Li, Wei Jin, Hongzhi Wen, Xiaochi Wei, Shuaiqiang Wang, Dawei Yin, Wenqi Fan, Hui Liu, and Jiliang Tang. 2023. Exploring the Potential of Large Language Models (LLMs) in Learning on Graphs. arXiv:2307.03393 [cs.LG]  https://arxiv.org/abs/2307.03393

[2] Yihong Dong, Kangcheng Luo, Xue Jiang, Zhi Jin, and Ge Li. 2023. PACE: Improving Prompt with Actor-Critic Editing for Large Language Model. (8 2023). http://arxiv.org/abs/2308.10088

[3] Yury Gorishniy, Ivan Rubachev, and Artem Babenko. 2022. On embeddings for numerical features in tabular deep learning. Advances in Neural Information Processing Systems 35 (2022), 24991–25004.

[4] Zellig S. Harris. 1954. Distributional Structure. WORD 10, 2-3 (1954), 146–162. https://doi.org/10.1080/00437956.1954.11659520 arXiv:https://doi.org/10.1080/00437956.1954.11659520

[5] Sihao Hu, Zhen Zhang, Bingqiao Luo, Shengliang Lu, Bingsheng He, and Ling Liu. 2023. BERT4ETH: A Pre-Trained Transformer for Ethereum Fraud Detection. In Proceedings of the ACM Web Conference 2023 (Austin, TX, USA) (WWW '23). Association for Computing Machinery, New York, NY, USA, 2189–2197. https://doi.org/10.1145/3543507.3583345

[6] Pingchuan Ma, Zongjie Li, Ao Sun, and Shuai Wang. 2023. "Oops, Did I Just Say That?" Testing and Repairing Unethical Suggestions of Large Language Models with Suggest-Critique-Reflect Process. (5 2023). http://arxiv.org/abs/2305.02626

[7] Tomas Mikolov, Kai Chen, Greg Corrado, and Jeffrey Dean. 2013. Efficient Estimation of Word Representations in Vector Space. arXiv:1301.3781 [cs.CL]

[8] The Hacker News. 2023. New AI Tool 'FraudGPT' Emerges, Tailored for Sophisticated Attacks. https://thehackernews.com/2023/07/new-ai-tool-fraudgpt-emerges-tailored.html

[9] Jack Nicholls, Aditya Kuppa, and Nhien-An Le-Khac. 2021. Financial Cybercrime: A Comprehensive Survey of Deep Learning Approaches to Tackle the Evolving Financial Crime Landscape. IEEE Access 9 (2021), 1–1. https://doi.org/10.1109/access.2021.3134076

[10] Jack Nicholls, Aditya Kuppa, and Nhien-An Le-Khac. 2023. SoK: The Next Phase of Identifying Illicit Activity in Bitcoin. 2023 IEEE International Conference on Blockchain and Cryptocurrency (ICBC), 1–10. https://doi.org/10.1109/ICBC56567.2023.10174963

[11] OpenAI. last accessed: 6th October 2023. OpenAI GPT-3 Models [text-davinci-003]. https://platform.openai.com/docs/models/gpt-3-5.

[12] XiPeng Qiu, TianXiang Sun, YiGe Xu, YunFan Shao, Ning Dai, and XuanJing Huang. 2020. Pre-trained models for natural language processing: A survey. Science China Technological Sciences 63, 10 (2020), 1872–1897. https://doi.org/10.1007/s11431-020-1647-3

[13] Alec Radford, Jong Wook Kim, Chris Hallacy, Aditya Ramesh, Gabriel Goh, Sandhini Agarwal, Girish Sastry, Amanda Askell, Pamela Mishkin, Jack Clark, Gretchen Krueger, and Ilya Sutskever. 2021. Learning Transferable Visual Models From Natural Language Supervision. In *Proceedings of the 38th International Conference on Machine Learning (Proceedings of Machine Learning Research, Vol. 139)*, Marina Meila and Tong Zhang (Eds.). PMLR, 8748–8763. https://proceedings.mlr.press/v139/radford21a.html

[14] Ashish Vaswani, Noam Shazeer, Niki Parmar, Jakob Uszkoreit, Llion Jones, Aidan N. Gomez, Lukasz Kaiser, and Illia Polosukhin. 2017. Attention Is All You Need. (6 2017). http://arxiv.org/abs/1706.03762

[15] Jingfeng Yang, Hongye Jin, Ruixiang Tang, Xiaotian Han, Qizhang Feng, Haoming Jiang, Bing Yin, and Xia Hu. 2023. Harnessing the Power of LLMs in Practice: A Survey on ChatGPT and Beyond. (4 2023). http://arxiv.org/abs/2304.13712

[16] Chaoning Zhang, Chenshuang Zhang, Sheng Zheng, Yu Qiao, Chenghao Li, Mengchun Zhang, Sumit Kumar Dam, Chu Myaet Thwal, Ye Lin Tun, Le Luang Huy, Donguk kim, Sung-Ho Bae, Lik-Hang Lee, Yang Yang, Heng Tao Shen, In So Kweon, and Choong Seon Hong. 2023. A Complete Survey on Generative AI (AIGC): Is ChatGPT from GPT-4 to GPT-5 All You Need? (3 2023). http://arxiv.org/abs/2303.11717

[17] Wayne Xin Zhao, Kun Zhou, Junyi Li, Tianyi Tang, Xiaolei Wang, Yupeng Hou, Yingqian Min, Beichen Zhang, Junjie Zhang, Zican Dong, Yifan Du, Chen Yang, Yushuo Chen, Zhipeng Chen, Jinhao Jiang, Ruiyang Ren, Yifan Li, Xinyu Tang, Zikang Liu, Peiyu Liu, Jian-Yun Nie, and Ji-Rong Wen. 2023. A Survey of Large Language Models. (3 2023). http://arxiv.org/abs/2303.18223